\pdfoutput=1

\documentclass[a4paper, 11pt]{article}

\usepackage{EACL2023}

\usepackage{times}
\usepackage{latexsym}
\usepackage[T1]{fontenc}
\usepackage[utf8]{inputenc}
\usepackage{microtype}
\usepackage{inconsolata}
\usepackage{tabularx}
\usepackage{graphicx}
\usepackage{amsmath}
\usepackage{caption}
\usepackage[titletoc,title]{appendix}
\usepackage{soul}
\usepackage{caption}
\usepackage{amsthm}        
\usepackage{amssymb}
\usepackage{makecell}
\usepackage{array,multirow}
\usepackage{booktabs}
\usepackage{CJKutf8}
\usepackage{makecell}
\RequirePackage{filecontents}
\usepackage[strict]{changepage}
\usepackage{subcaption}
\usepackage{etoolbox}
\usepackage[ruled,vlined]{algorithm2e}

\title{Train Global, Tailor Local: \\
Minimalist Multilingual Translation  \\
into Endangered Languages 
}

\author{
  Zhong Zhou\\
  {\small Carnegie Mellon University}\\
  {\small \tt zhongzhou@cmu.edu}
  \\\And
  Jan Niehues\\
  {\small Karlsruhe Institute of Technology}\\
  {\small \tt jan.niehues@kit.edu}\\
  \\\And
  Alex Waibel\\
  {\small Carnegie Mellon University}\\
  {\small Karlsruhe Institute of Technology}\\
  {\small \tt alex@waibel.com }
}

\begin{document}
\maketitle

\begin{abstract}
In many humanitarian scenarios, translation into severely low resource languages often does not require a universal translation engine, but a dedicated \textit{text}-\textit{specific} translation engine. For example, healthcare records, hygienic procedures, government communication, emergency procedures and religious texts are all limited texts. While generic translation engines for all languages do not exist, translation of multilingually known limited texts into new, endangered languages may be possible and reduce human translation effort. We attempt to leverage translation resources from many rich resource languages to efficiently produce best possible translation quality for a well \textit{known text}, which is  available in multiple languages, in a new, severely low resource language. We examine two approaches: 1.) best selection of seed sentences to jump start translations in a new language in view of best generalization to the remainder of a larger targeted text(s), and 2.) we adapt large general multilingual translation engines from many other languages to focus on a specific text in a new, unknown language. We find that adapting large pretrained multilingual models to the domain/text first and then to the severely low resource language works best. If we also select a best set of seed sentences, we can improve average chrF performance on new test languages from a baseline of 21.9 to 50.7, while reducing the number of seed sentences to only $\sim$1,000 in the new, unknown language.

\end{abstract}

\section{Introduction}
A language dies when no one speaks it. An endangered language is a
language that is spoken by enough people that it could survive under favorable
conditions but few or no children are learning it
\citep{crystal2002language, kincade1991decline, wurm2001atlas}. 
More than half of the ~7,139 languages will die in the next 80 years \citep{austin2011cambridge, eberhard2021ethnologue}. 
Endangered languages may survive and thrive if they gain prestige, power and visibility \citep{crystal2002language}. 
Frisian, for example, struggles to gain prestige in Germany, and is endangered even though it has a large number of speakers. 
Hebrew, conversely, has been revived as a spoken language because it is critical to the development and identity of the Jewish community. We empower endangered language communities by exercising a language.  This can be achieved by translating important texts to their language so that these communities can gain information, knowledge, power and visibility in their own language. One life-saving example of this knowledge-transfer is translating water, sanitation
and hygiene (WASH) text into their languages, a process that has long started before the COVID-19 pandemic but has gained much attention since then \citep{thampi2020s, reddy2017water}.
\begin{figure}
  \centering
  \includegraphics[width=1.0\linewidth]{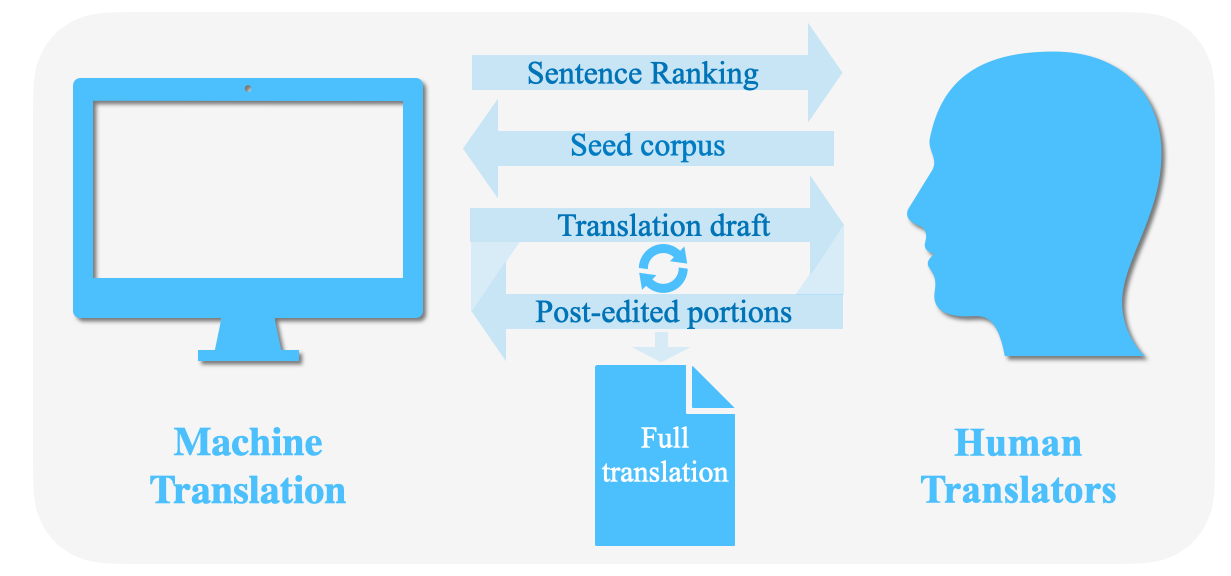}
  \caption{Translation workflow for endangered languages.}
  \label{fig:alworflow}
\end{figure}

The problem in these scenarios, therefore, is not to build a high accuracy  translation engine for \textit{any texts }using huge data corpora, but rather to build a good translation for a \textit{known} text (for which translations in many other languages exist), but in a new language with only extremely little seed data (a few hundred sentences).  We assume there is little to no endangered language data and few human translators. To produce high quality translation, existing methods rely on a seed corpus produced by human translators. Previous work has shown progress in using extremely small seed corpora with as small as $\sim$1,000 lines of data and has found that random sampling performs better than choosing a fixed portion of the text to build a seed corpus \citep{zhou2021family, lin2020pre, qi2018and}.
But
researchers have yet to 1.) examine various Active Learning (AL) methods to improve accuracy and effectiveness in building better optimized seed corpora so as to minimize the initial human effort and 2.) completely solve the problem of using large multilingual models for representational learning so that we can train (or adapt) them to a new language using an extremely small seed corpus.

To solve these two problems, we propose explainable and robust active learning methods that perform as well as or better than random sampling; we transfer methods learned on data of known languages to the new, endangered language. We also examine different training schedules and we find a strategic way of growing large multilingual models in a multilingual and multi-stage fashion with extremely small endangered seed corpora. 

In our translation workflow, human translators are informed by machine sentence ranking to produce a seed corpus. Machine systems then use this seed corpus to produce a full translation draft. Human translators post-edit the draft, and feed new data to machines each time they finish post-editing a portion of the text. In each iteration, machines produce better and better drafts with new data, and human translators find it easier and faster to post-edit. Together they complete the translation of the whole text into an endangered 
language (Figure~\ref{fig:alworflow}).  

\begin{figure}
  \centering
  \includegraphics[width=1.0\linewidth]{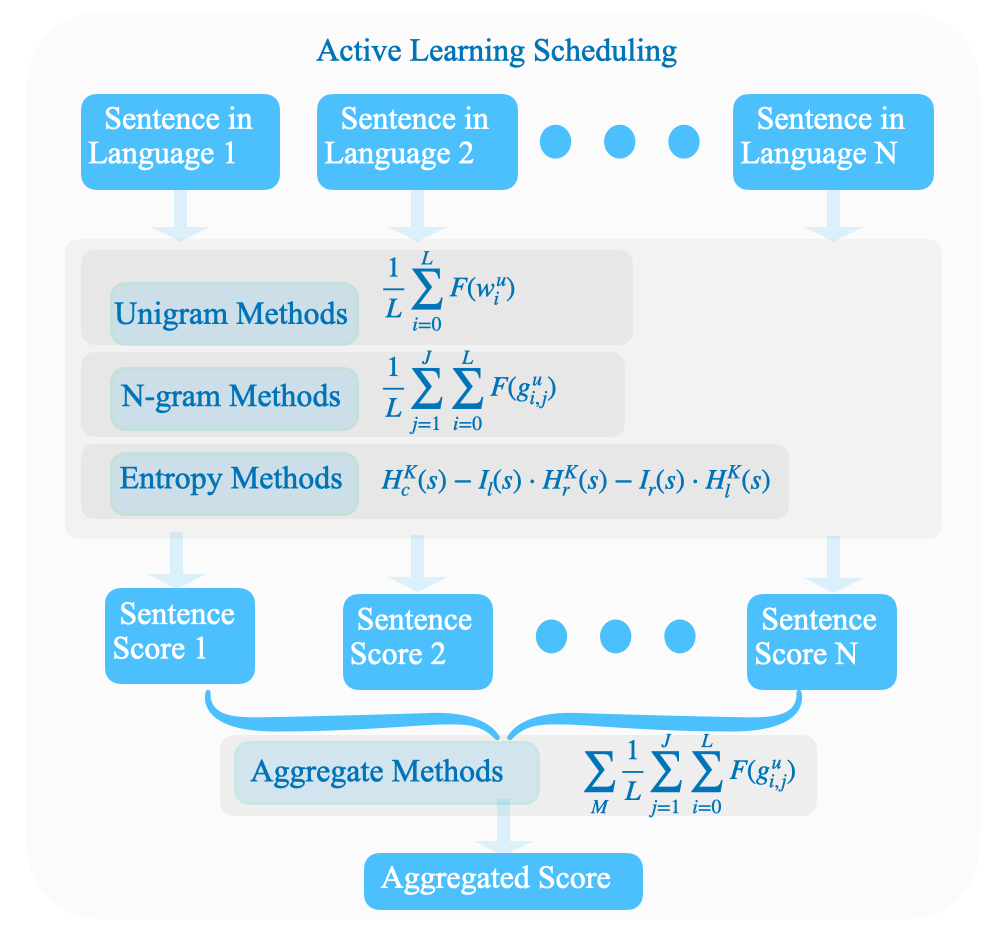}
  \caption{Visualizing different active learning methods. We score and rank each sentence in a text corpus. }
  \label{fig:active}
\end{figure}

To produce sentence ranking, traditional active learning approaches assume abundant data, but we have little to no data in the target endangered language. We question this assumption and build seed corpora by ranking 
all sentences in existing translations from other languages to generalize to a new, endangered language. This ranking is target-independent as we do not require any endangered language data. To produce such a ranking, we explore active learning methods (Table~\ref{table:ngramScore}). For each reference language, we build unigram, n-gram and entropy models (Figure~\ref{fig:active}). To prevent any language from overpowering the ranking, we aggregate sentence scores across multiple languages and rank the final aggregation. To select the pool of languages for aggregation, we build methods on different voting mechanisms. 

To curate a seed corpus in the new, endangered language where we have no data initially, we pass the sentence ranking learned from known languages to human translators. Human translators take this ranking, and translate the top few ($\sim$1,000 or less) sentences, curating the seed corpus. 

To train on such small seed corpus, we find pretraining to be key. For the pretrained model, we either create our own pretrained model by training on known languages, or use an existing pretrained model. We explore both paths in our work, with and without activating the knowledge in existing large pretrained models. We observe an average increase of 28.8 in chrF score over the baselines. 

Our contribution is three-fold: 1. We develop 14 active learning methods on known languages and transfer ranking to the new, endangered language; 2. 
We activate the knowledge of large multilingual models by proposing multilingual and multi-stage adaptations through 24 different training schedules; we find that adapting pretrained models to the domain and then to the endangered language works best; 
3. We aggregate scores from 115 languages to provide a universal ranking and increase robustness by \textit{relaxed memoization} method.

\section{Related Works} \label{relatedwork}
\subsection{Translation into Endangered Languages}
Recent advances have succeeded in building multilingual
methods to translate from multiple rich resource
languages to a new, endangered language
\citep{johnson2017google, ha2016toward, firat2016multi, zhou2018massively, zhou2018paraphrases}.
Many have demonstrated good transfer learning to low resource languages
\citep{zhou2021family, lin2020pre, qi2018and}, while some work on zero-shot learning
\citep{neubig2018rapid, pham2019improving, philip2020monolingual, karakanta2018neural, zhang2020improving, chen2022towards, chen2021zero}. 
However, zero-shot learning is volatile
and unstable, 
so 
we choose to use extremely small data instead. 

\subsection{Active Learning in Machine Translation}
Active learning has a long history in machine translation
\citep{settles2012active, eck2005low, gonzalez2012active}.
Random sampling is often surprisingly powerful
\citep{kendall1938randomness, knuth19913, sennrich2015improving}.
There is extensive research to beat random sampling by methods based on 
entropy \citep{koneru2022cost}, coverage and
uncertainty \citep{peris2018active, zhao2020active},
clustering \citep{haffari2009activea, gangadharaiah2009active},
consensus \citep{haffari2009activeb}, syntactic parsing
\citep{miura2016selecting}, density and diversity
\citep{koneru2022cost, ambati2011multi}, and learning to learn active learning strategies \citep{liu2018learning}.

\subsection{Large Pretrained Multilingual Model} 
The state-of-the-art multilingual machine translation systems translate from many source languages to many target languages 
\cite{johnson2017google, ha2016toward, zoph2016multi}. The bottleneck in building such systems is in computation limits, as the training data increases quadratically with the number of languages. Some companies have built and released large pretrained multilingual models
\citep{liu2020multilingual, tang2020multilingual}.
M2M100 is trained in 100 languages
\citep{fan2021beyond, schwenk2019ccmatrix, el2019massive} and covers a few endangered languages.

\section{Methods}
We translate a fixed text that is available in many languages to a new, endangered language. In our translation workflow, we first develop active learning methods to transfer sentence ranking from known languages to a new, endangered language. We then pass this ranking to human translators for them to translate the top few ($\sim$1,000 or less) sentences into the endangered language, curating the seed corpus. We finally train on the seed corpus, either from scratch or from a pretrained model. 

We build training schedules on an extremely small seed corpus, we also build active learning strategies of creating and transferring the sentence ranking to the new, endangered language. We propose and compare 24 training schedules and 14 active learning methods for machine translation into a new, endangered language. To compare all active learning algorithms fairly, we use the same translation system unit as a control for all experiments, varying only the seed corpora built by different methods. We select the same number of words in all seed corpora as most translators are paid by the number of words
\citep{bloodgood2014bucking, eck2008developing, tomanek2009semi}.

\begin{figure}
  \centering
  \includegraphics[width=1.0\linewidth]{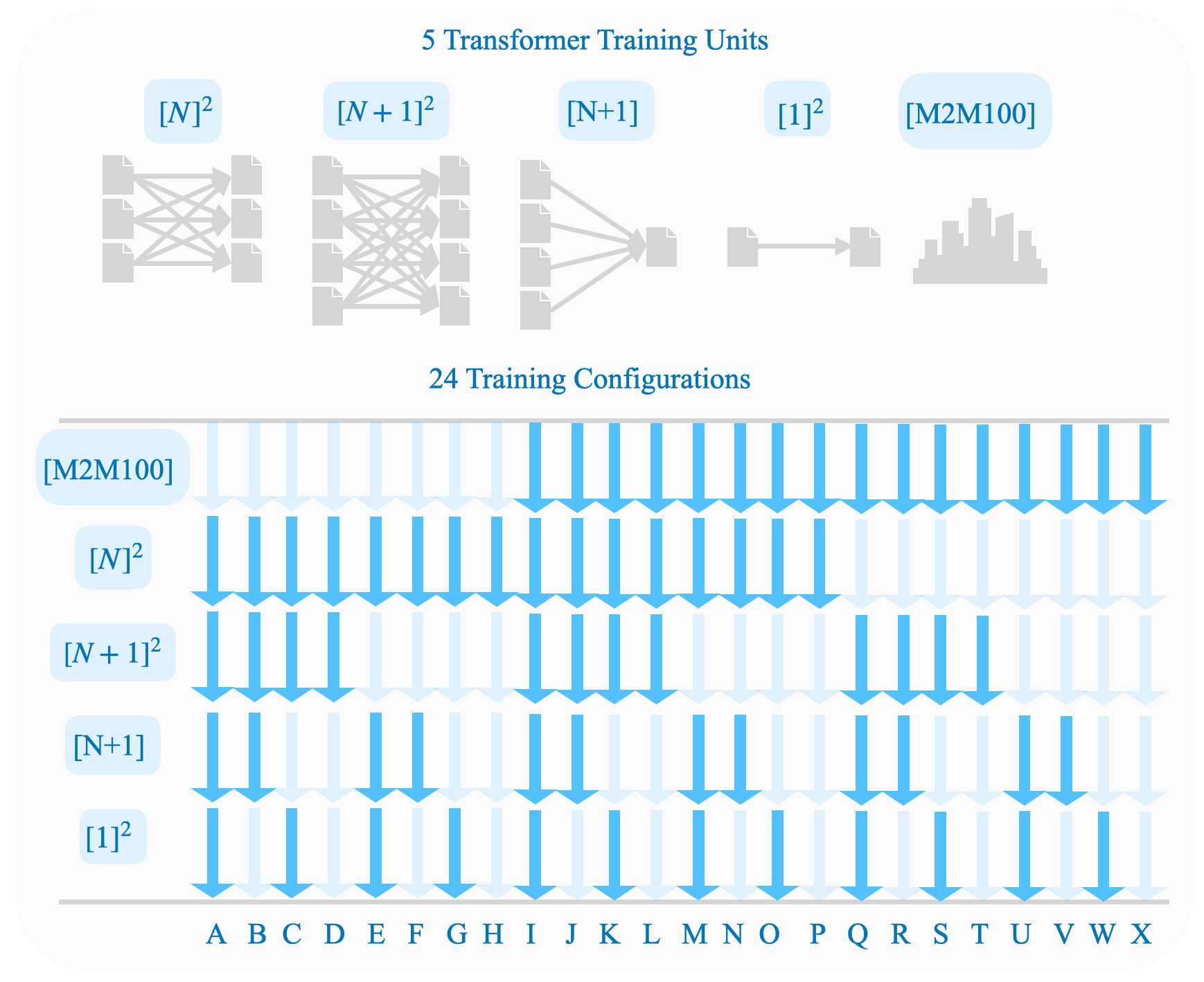}
  \caption{24 different training schedules.\\
  {[N]}: multilingual model on N neighboring languages\\
  {[N+1]}$^2$: multi-target model with endangered language \\
  {[N+1]}: single-target model with endangered language\\
  {[1]}$^2$: autoencoder in endangered language.}
  \label{fig:config13}
\end{figure}

\subsection{Training Schedules}
In our setup we have the new, endangered language as the target language, and we have a few neighboring languages as the source languages that are either in the same linguistic language family or geographically close to facilitate linguistic transfer. In effect, we have $N$ source languages with full translations of the text and a new and endangered language that has an extremely small seed corpus. 

We use the state-of-the-art multilingual transformer prepending both source and target language labels to each source sentence \citep{johnson2017google, ha2016toward}. For precise translation for all named entities, we use an existing method of \textit{order-preserving named entity translation} by masking each named entity with ordered \texttt{\_\_NE}s using a parallel multilingual lexicon table in 125 languages \citep{zhou2021family, wu2018creating}. 

Using this multilingual transformer architecture as a base, we build 5 training units on the small seed corpus of the new, endangered language and the existing translations of known languages. 
We let {[N]}$^2$ denote the training of all source languages in a N-by-N multilingual transformer. We let {[N+1]}$^2$ denote the training of all languages including the endangered language in a (N+1)-by-(N+1) multilingual transformer. We let {[N+1]} denote the (N+1)-by-1 multilingual transformer that focuses on translating into the endangered language. We let {[1]}$^2$ be the autoencoder on the endangered language. 

Our translation system is built on these 5 training units: an optional {[M2M100]} \citep{fan2021beyond}, {[N]}$^2$, {[N+1]}$^2$, {[N+1]} and {[1]}$^2$. 
These 5 stages increase in specificity while they decrease in data size. Building on them, we show 24 different training schedules, among which 8 are pretrained with in-domain data and 16 are pretrained with out-of-domain large multilingual models (Figure~\ref{fig:config13}). We only consider models with pretraining and therefore do not exhaust all 32 training schedules. 

\subsection{Active Learning Strategies} \label{2023methods}
We have two baselines: the linguistic baseline of the excerpt-based approach, \textit{Luke}, and the statistical baseline of random sampling, \textit{Rand}. The excerpt-based approach, which selects a portion of the text with consecutive sentences, preserves the text's formality, cohesion and context but lacks global coverage. Random sampling increases global coverage but sacrifices local coherence.

\subsubsection{N-gram Approach}
Many researchers count the number of unknown n-grams as score functions to solve the knapsack problem, covering all vocabulary \citep{eck2008developing, eck2005low, haffari2009activea}.
Instead of solving the knapsack problem,  we choose sentences to partially cover the vocabulary and build an extremely small seed corpus. To cover the vocabulary strategically, we sum the frequency counts of the unknown n-grams to increase density. These frequency counts promote frequent words for learning to be meaningful in the extremely low resource scenario. In Table~\ref{table:ngramScore} we denote frequency function by $F(\cdot)$, denote sequence length by $L$ and denote the highest n-gram order by $J$. 

\begin{table}[t]
  \small
  \centering
  \begin{tabularx}{\columnwidth}{p{0.7cm}p{2.5cm}p{3.4cm}}
    \toprule
    Name & Description & Score Function \\ 
    \midrule 
    \textit{S} & Frequency sum of unknown words & $ \sum\limits_{i=0}^L F(w^{u}_i)$ \\
    \textit{SN} & Normalized \textit{S} by $L$ & $\frac{1}{L} \sum\limits_{i=0}^L F(w^{u}_i)$ \\
    \textit{SNG$_{J}$} & Normalized Frequency sum of n-grams up to $J$ & $\frac{1}{L} \sum\limits_{j=1}^J \sum\limits_{i=0}^L F(g^{u}_{i,j})$ \\
    \textit{AGG$^{M}_{J}$} & Aggregation of n-gram scores up to $J$ with set $M$ & $\sum\limits_{M} \frac{1}{L} \sum\limits_{j=1}^J \sum\limits_{i=0}^L F(g^{u}_{i,j})$ \\
    \textit{ENT$^{K}$} & Entropy methods, $K$ is KenLM or not & $H_{c}^K(s) - I_{l}(s) \cdot H_{r}^K(s) - I_{r}(s) \cdot H_{l}^K(s)$ \\
    \bottomrule
  \end{tabularx}
    \caption{Summary of score functions. 
    }
\label{table:ngramScore}
\end{table}

\begin{table*}[t]
  \small
  \centering
  \begin{tabularx}{\textwidth}{p{0.9cm}p{0.02cm}p{1.32cm}p{12.5cm}}
    \toprule 
    Target & L & Family & Source Languages \\
    \midrule
    Frisian & 0 & Germanic & English*, German, Dutch, Norwegian, Afrikaans, Swedish, French, Italian, Portuguese, Romanian \\  
    Hmong & 0 & Hmong–Mien & Komrem*, Vietnamese, Thai, Chinese, Myanmar, Haka, Tangsa, Zokam, Siyin, Falam \\
    Pokomchi & 0 & Mayan & Chuj*, Cakchiquel, Mam, Kanjobal, Cuzco, Ayacucho, Bolivian, Huallaga, Aymara, Guajajara \\
    Turkmen & 1 & Turkic & Kyrgyz*, Tuvan, Uzbek, Karakalpak, Kazakh, Azerbaijani, Japanese, Korean, Finnish, Hungarian \\
    Sesotho & 1 & Niger–Congo & Yoruba*, Gikuyu, Xhosa, Kuanyama, Kpelle, Fon, Bulu, Swati, Venda, Lenje \\
    Welsh & 1 & Celtic & English*, German, Danish, Dutch, Norwegian, Swedish, French, Italian, Portuguese, Romanian \\
    Xhosa & 2 & Nguni & Swati*, Gikuyu, Sesotho, Yoruba, Lenje, Gbaya, Afrikaans, Wolaitta, Kuanyama, Bulu \\
    Indonesian & 3 & Austronesian & Javanese*, Malagsy, Tagalog, Ilokano, Cebuano, Fijian, Sunda, Zokam, Wa, Maori \\	
    Hungarian & 4 & Uralic & Finnish*, French, English, German, Latin, Romanian, Swedish, Spanish, Italian, Portuguese \\
    Spanish & 5 & Romance & English*, German, Danish, Dutch, Norwegian, Swedish, French, Italian, Portuguese, Romanian \\
    \bottomrule
  \end{tabularx}
    \caption{Summary of different target languages used \citep{campbell2018cataloguing, collin2010ethnologue}.
    L, resource level, is from a scale of 0 to 5 \citep{joshi2020state}.
    Reference languages used for active learning methods except aggregate methods are starred. }
\label{table:ethnologue_languages}
\end{table*}

\subsubsection{Entropy Approach}
Many have worked on entropy methods in modelling density and diversity
\citep{ambati2011multi, eck2008developing, zeng2019empirical, haffari2009activea}.
We use traditional Language Models (LMs) instead of neural language models, as our data size is extremely small. 
For implementations of LMs, we use KenLM and NLTK's LM because of their simplicity and speed, especially KenLM \citep{heafield2011kenlm, loper2002nltk}. 
In Table~\ref{table:ngramScore} we let $H(\cdot)$ be the cross entropy function, with the choice of KenLM (K) or NLTK (N). To separate training from testing in using language models, we divide the data into three portions, the sentences that we have chosen (\textit{c}), and the remaining that are split equally into two parts, left 
(\textit{l}) and right (\textit{r}). Let $I_{l}(\cdot)$ and $I_{r}(\cdot)$ be indicator functions to show whether a sentence belongs to the left or the right. We aim to maximize the diversity $H_{c}$ and optimize density by adjusting $H_{l}$ and $H_{r}$ \citep{koneru2022cost}. 

\subsubsection{Aggregation Approach}
To prevent any language from overpowering the ranking, we aggregate sentence scores across different languages (Figure~\ref{fig:active}). We investigate the use of a customized set of languages for each endangered language, versus the use of a universal set of languages representing world languages. The former requires some understanding of the neighboring languages, the latter requires careful choices of the representative set \citep{blasi2021systematic}.

We have 4 aggregation methods: \textit{one-vote-per-language} (L), where we aggregate over all languages, \textit{one-vote-per-family} (F), where we aggregate over languages representing the top few families, \textit{one-vote-per-person} (P), where we aggregate over the top few most spoken languages, and \textit{one-vote-per-neighbor} (N), where we aggregate over a customized set of neighboring languages. For the world language distribution, L covers all, F samples across it, P covers the head, while N creates a niche area around the endangered language.

Aggregation decreases variance and increases accuracy. Typical aggregation 
involve taking the sum or the average. Since they have the same
effect on sentence ranking, we take the sum for simplicity.

To save space and time, we devise \textit{relaxed memoization}. 
At every step, we compute sentence score for each language, producing a score matrix of languages versus sentences. We update entries that are affected by the selected sentence, cache and reuse other entries. Further parallelism  
results in >360 times speedup, from $\sim$6.5 months to $\sim$13 hours. 

\begin{table*}[t]
  \small
  \centering
  \begin{tabularx}{\textwidth}{p{1.8cm}p{0.8cm}p{0.8cm}p{1.0cm}p{0.8cm}p{0.8cm}p{0.6cm}p{0.6cm}p{1.0cm}p{1.0cm}p{0.8cm}p{0.8cm}} 
   \toprule
   $\uparrow$chrF & Frisian & Hmong & Pokomchi & Turkmen & Sesotho & Welsh & Xhosa & Indonesian & Hungarian & Spanish & Average \\
    \midrule
    \multicolumn{12}{l}{\textbf{Baselines:} }\\
    + Bilingual & 23.1 & 25.0 & 28.7 & 18.9 & 25.2 & 22.2 & 21.4 & 27.2 & 20.1 & 22.1 & 23.4 \\
    + Multilingual & 28.0 & 28.1 & 31.9 & 22.6 & 28.3 & 26.5 & 23.9 & 29.7 & 22.3 & 26.8 & 26.8 \\
    \midrule
    \multicolumn{12}{l}{\textbf{Our Models:} }\\
    + Schedule B & 50.5 & 43.9 & 42.8 & 38.9 & 43.2 & 46.0 & 34.9 & 47.2 & 37.4 & 50.1 & 43.5 \\
    + Active (AL) & 53.6 & 45.7 & 44.4 & 40.3 & 44.9 & 47.7 & 36.8 & 49.1 & 39.0 & 52.7 & 45.4 \\
    \bottomrule
  \end{tabularx}
  \caption{Results for translation into 10 languages that are 
  new and severely low resourced to the system, independent of M2M100. }
\label{table:4by10}
\end{table*}

\begin{table}[t]
  \small
  \centering
  \begin{tabularx}{\columnwidth}{p{1.8cm}p{0.5cm}p{0.5cm}p{0.95cm}p{0.7cm}p{0.7cm}} 
   \toprule
   $\uparrow$chrF & Frisian & Welsh & Hungarian & Spanish  & Average \\
    \midrule
    \multicolumn{6}{l}{\textbf{Baselines:} }\\
    + Bilingual & 23.1 & 22.2 & 20.1 & 22.1 & 21.9 \\
    + Multilingual & 28.0 & 26.5 & 22.3 & 26.8 & 25.9 \\
    + M2M100 & 26.0 & 9.9 & 38.8 & 47.5 & 24.9\\ 
    \midrule
    \multicolumn{6}{l}{\textbf{Our Models:} }\\
    + Schedule I & 53.5 & 49.5 & 42.2 & 53.2 & 49.6 \\ 
    + Active (AL)  & 54.9 & 49.8 & 43.2 & 54.9 & 50.7 \\
    \bottomrule
  \end{tabularx}
  \caption{Results for translation into 4 languages that are 
  new and severely low resourced to the system, 
  activating knowledge in M2M100 and leveraging active learning.    
  }
\label{table:5by10}
\end{table}

\begin{table*}[t]
  \small
  \centering
  \begin{tabularx}{\textwidth}{p{1cm}XXXXXXXXXXXXXXXX} 
    \toprule
     Network & I & J & K & L & M & N & O & P & Q & R & S & T & U & V & W & X \\
    \midrule
    {[M2M100]} & $\Downarrow$ & $\Downarrow$ & $\Downarrow$ & $\Downarrow$  & $\Downarrow$ & $\Downarrow$ & $\Downarrow$ & $\Downarrow$  & $\Downarrow$ & $\Downarrow$ & $\Downarrow$ & $\Downarrow$  & $\Downarrow$ & $\Downarrow$ & $\Downarrow$ & $\Downarrow$ \\
    {[N]}$^2$  & $\Downarrow$ & $\Downarrow$ & $\Downarrow$ & $\Downarrow$  & $\Downarrow$ & $\Downarrow$ & $\Downarrow$ & $\Downarrow$  &  &  &  & &  &  &  \\
    {[N+1]}$^2$  & $\Downarrow$ & $\Downarrow$ & $\Downarrow$ & $\Downarrow$ &  &  &  &   & $\Downarrow$ & $\Downarrow$ & $\Downarrow$ & $\Downarrow$ &  &  &  \\ 
    {[N+1]} & $\Downarrow$ & $\Downarrow$ & & & $\Downarrow$ & $\Downarrow$ & & & $\Downarrow$ & $\Downarrow$ & & & $\Downarrow$ & $\Downarrow$ &\\
    {[1]}$^2$  & $\Downarrow$ & & $\Downarrow$ & & $\Downarrow$ & &  $\Downarrow$ & & $\Downarrow$ & & $\Downarrow$ & & $\Downarrow$ & & $\Downarrow$ &\\
    \midrule
    $\uparrow$chrF & \bf{52.9} & \bf{51.8} & 49.5 & \bf{52.8} & \bf{52.7} & \bf{51.9} & 27.4 & 16.9 & 49.6 & 48.5 & 39.6 & 48.7 & 48.5 & 45.7 & 27.8 & 26.3 \\
    $\downarrow$cTER & 0.492 & 0.508 & 0.482 & 0.488 & 0.493 & 0.502 & 0.654 & 0.800 & 0.530 & 0.546 & 0.553 & 0.539 & 0.538 & 0.579 & 0.650 & 0.667 \\
    $\uparrow$BLEU & 28.8 & 27.9 & 24.2 & 28.9 & 28.8 & 28.2 & 3.0 & 0.6 & 24.8 & 24.2 & 13.9 & 24.3 & 24.5 & 22.0 & 3.4 & 3.3 \\ 
    $\uparrow$COMET & \text{-}0.56 & \text{-}0.59 & \text{-}0.63 & \text{-}0.53 & \text{-}0.56  & \text{-}0.57 & \text{-}1.28 & \text{-}1.75 & \text{-}0.67  & \text{-}0.70 & \text{-}0.89 & \text{-}0.68 & \text{-}0.69  & \text{-}0.80 & \text{-}1.21 & \text{-}1.30 \\
    $\uparrow$BERTS &  0.891 & 0.889 & 0.886 & 0.892 & 0.891 & 0.890 & 0.813 & 0.775 & 0.883 & 0.881 & 0.861 & 0.882 & 0.880 & 0.873 & 0.823 & 0.819 \\
    \bottomrule
\end{tabularx}
    \caption{Comparing 16 training schedules with M2M100. BERTS is BERTScore, cTER is characTER and LRatio is length ratio.  
    }
\label{table:config2}
\end{table*}

\begin{table}[t]
  \small
  \centering
  \begin{tabularx}{\columnwidth}{p{.8cm}XXXXXXXX} 
    \toprule
    Network & A & B & C & D & E & F & G & H \\
    \midrule
    {[N]}$^2$  & $\Downarrow$ & $\Downarrow$ & $\Downarrow$ & $\Downarrow$  & $\Downarrow$ & $\Downarrow$ & $\Downarrow$ & $\Downarrow$ \\
    {[N+1]}$^2$  & $\Downarrow$ & $\Downarrow$ & $\Downarrow$ & $\Downarrow$ &  &  &   \\ 
    {[N+1]} & $\Downarrow$ & $\Downarrow$ & & & $\Downarrow$ & $\Downarrow$ & \\
    {[1]} $^2$  & $\Downarrow$ &  & $\Downarrow$ &  & $\Downarrow$ & &  $\Downarrow$ &  \\ 
    \midrule
    $\uparrow$chrF & 38.7 & \bf{51.1} & 35.6 & 50.8 & 43.4 & \bf{51.2} & 25.6 & 24.1 \\
    $\downarrow$cTER & 0.555 & 0.517 & 0.572 & 0.515 & 0.523 & 0.507 & 0.650 & 0.682  \\
    $\uparrow$BLEU & 12.5 & 24.9 & 9.2 & 24.5 & 17.5 & 26.2 & 2.5 & 2.1 \\ 
    $\uparrow$COMET & \text{-}0.87 & \text{-}0.66 & \text{-}0.91 & \text{-}0.65 & \text{-}0.81 & \text{-}0.63 & \text{-}0.99 & \text{-}1.02 \\
    $\uparrow$BERTS & 0.850 & 0.882 & 0.839 & 0.884 & 0.865 & 0.885 & 0.801 & 0.794 \\
    \bottomrule
\end{tabularx}
    \caption{Comparing 8 training schedules without M2M100. \\
     {[N]}$^2$: multilingual model on N neighboring languages\\
     {[N+1]}$^2$ : multi-target model with endangered language \\
     {[N+1]}: single-target model with endangered language\\
     {[1]}$^2$: autoencoder in endangered language.  
    }
\label{table:config1}
\end{table}

\subsection{Evaluation Method and Metrics}
Existing multilingual systems produce multiple outputs from all source languages, rendering comparison messy. To simplify, we combine translations from all source languages into one by an existing \textit{centeredness method} \citep{zhou2021family}. Using this method, we score each translated sentence by the sum of its similarity scores to all others. We rank these scores and take the highest score as our combined score. The expected value of the combined score is higher than that of each source. 

To compare effectively, we control all test sets to be the same. Since different active learning strategies produce different seed corpora to be used as training and validation sets, the training and validation sets vary. Their complement, the test sets therefore also vary, rendering comparison difficult. To build the same test set, we devise an \textit{intersection method}. We take the whole text and carve out all seed corpora, that is, all training and validation sets from all experiments. The remaining is the final test set, which is the intersection of all test sets. 

Our metrics are: chrF, characTER, BLEU, COMET score, and BERTscore \citep{popovic2015chrf, wang2016character, post-2018-call, zhang2019bertscore, stewart-etal-2020-comet, rei2021mt}. We prioritize chrF over BLEU for better accuracy, fluency and expressive power in morphologically-rich languages \citep{papineni2002bleu}.

\section{Data}
Existing research classifies world languages into Resource 0 to 5, with 0 having the lowest resource and 5 having the highest  \citep{joshi2020state}. We choose 10 target languages ranging from Resource 0 to 5 (Table~\ref{table:ethnologue_languages}). For each target language we choose ten neighboring languages as source
languages (Table~\ref{table:ethnologue_languages}). 
We prioritize Resource 0 to 2 languages as real endangered languages, and we use Resource 3 to 5 languages as hypothetical ones. 

To translate into these languages, our text is the Bible in 125 languages \citep{mayer2014creating}. Each endangered seed corpus contains $\sim$3\% of the text, while all other languages have full text. Our goal is to translate the rest of the text into the endangered language. In pretraining, we use a 80/10/10 split for training, validation and testing, respectively. In training, we use approximately a 3.0/0.2/96.8 split for training, validation and testing, respectively. Our training data for each experiment is $\sim$1,000 lines. We use BPE with size of $\sim$3,000 for the endangered language and $\sim$9,000 for the combined \citep{sennrich2016neural}. 

\begin{table*}[t]
  \small
  \centering
  \begin{tabularx}{\textwidth}{p{1.2cm}p{0.8cm}p{0.8cm}p{1.1cm}p{0.9cm}p{0.9cm}p{0.7cm}p{0.7cm}p{1.1cm}p{1.1cm}p{0.8cm}p{0.9cm}} 
   \toprule
   $\uparrow$chrF & Frisian & Hmong & Pokomchi & Turkmen & Sesotho & Welsh & Xhosa & Indonesian & Hungarian & Spanish & Average \\
    \midrule
    \multicolumn{12}{l}{\textbf{Baselines:} }\\
    + \textit{Luke} & 47.5 & 41.6 & 39.4 & 34.9 & 41.2 & 41.2 & 32.0 & 43.3 & 34.4 & 46.7 & 40.2\\
    + \textit{Rand} & 50.5 & 43.9 & 42.8 & 38.9 & 43.2 & 46.0 & 34.9 & 47.2 & 37.4 & 50.1 & 43.5 \\
    \midrule 
    \multicolumn{12}{l}{\textbf{Our Models:} }\\
    + \textit{S} & 49.2 & 38.5 & 40.4 & 35.2 & 39.0 & 41.9 & 32.5 & 43.5 & 35.1 & 48.0 & 40.3 \\
    + \textit{SN} & 50.9 & 43.9 & 43.2 & 38.3 & 41.6 & 43.2 & 36.1 & 46.9 & 36.7 & 50.3 & 43.1 \\
    + \textit{SNG$_{2}$} & 53.2 & \bf{46.1} & 43.3 & 39.5 & 44.4 & 45.8 & 36.6 & 48.4 & 37.8 & 51.8 & 44.7 \\
    + \textit{SNG$_{3}$} & 52.7 & 46.0 & \bf{44.5} & 39.6 & \bf{45.5} & 47.5 & \bf{36.8} & 48.9 & \bf{39.2} & 52.3 & 45.3 \\
    + \textit{SNG$_{4}$} & \bf{53.6} & 45.7 & 44.4 & \bf{40.3} & 44.9 & 47.7 & 36.8 & \bf{49.1} & 39.0 & \bf{52.7} & \bf{45.4} \\
    + \textit{SNG$_{5}$} & 53.0 & 45.6 & 43.9 & 39.7 & 45.4 & 46.7 & 36.8 & 49.1 & 38.4 & 52.5 & 45.1 \\
    + \textit{ENT$^{N}$} & 50.9 & 43.7 & 38.1 & 37.2 & 42.5 & 44.5 & 34.7 & 46.7 & 36.0 & 49.9 & 42.4 \\
    + \textit{ENT$^{K}$} & 52.7 & 45.7 & 43.5 & 40.2 & 44.6 & 45.2 & 36.4 & 49.0 & 39.1 & 51.8 & 44.8 \\
    + \textit{AGG$^{L}_{5}$} & 47.1 & 41.5 & 39.8 & 34.0 & 39.9 & 42.1 & 31.4 & 43.5 & 33.7 & 45.2 & 39.8 \\
    + \textit{AGG$^{F}_{5}$} & 45.0 & 38.4 & 38.5 & 32.4 & 38.8 & 47.1 & 30.4 & 41.2 & 33.3 & 44.2 & 38.9 \\
    + \textit{AGG$^{P}_{5}$} & 45.5 & 38.8 & 38.0 & 32.0 & 38.8 & \bf{48.2} & 30.5 & 41.0 & 33.2 & 44.0 & 39.0 \\
    + \textit{AGG$^{N}_{5}$} & 45.4 & 39.1 & 38.3 & 32.4 & 38.8 & 48.0 & 30.7 & 41.2 & 33.2 & 44.3 & 39.1 \\
    \bottomrule
  \end{tabularx}
  \caption{140 experiments comparing 14 active learning methods translating into 10 different languages with Schedule \textit{B}.}
\label{table:14by10}
\end{table*}

\begin{table}[t]
  \small
  \centering
  \begin{tabularx}{\columnwidth}{p{1.2cm}p{0.7cm}p{0.7cm}p{1.1cm}p{0.7cm}p{0.8cm}} 
    \toprule
    $\uparrow$chrF & Frisian & Welsh & Hungarian & Spanish  & Average \\
    \midrule
    \multicolumn{6}{l}{\textbf{Baselines:} }\\
    + \textit{Luke} & 49.3 & 44.3 & 38.8 & 48.4 & 45.2 \\
    + \textit{Rand} & 53.5 & 49.5 & 42.2 & 53.2 & 49.6 \\
    \midrule 
    \multicolumn{6}{l}{\textbf{Our Models:} }\\
    + \textit{S} & 51.9 & 45.9 & 40.4 & 51.1 & 47.3 \\
    + \textit{SN} & 54.8 & 47.4 & 42.3 & 53.2 & 49.4 \\
    + \textit{SNG$_{2}$} & 54.5 & 49.5 & 43.5 & 54.2 & 50.4 \\
    + \textit{SNG$_{3}$} & 54.4 & 50.4 & \bf{43.9} & 54.5 & \bf{50.8} \\
    + \textit{SNG$_{4}$} & \bf{54.9} & 49.8 & 43.2 & \bf{54.9} & 50.7 \\
    + \textit{SNG$_{5}$} & 54.5 & 50.1 & 43.5 & 54.1 & 50.6 \\
    + \textit{ENT$^{N}$} & 52.7 & 47.2 & 40.9 & 52.9 & 48.4 \\
    + \textit{ENT$^{K}$} & 54.6 & 49.4 & 43.5 & 53.8 & 50.3 \\
    + \textit{AGG$^{A}_{5}$} & 49.4 & 44.2 & 37.3 & 48.2 & 44.8\\
    + \textit{AGG$^{S}_{5}$} & 46.5 & 49.8 & 36.4 & 46.4 & 44.8\\
    + \textit{AGG$^{M}_{5}$} & 48.6 & 50.4 & 36.5 & 46.9 & 45.6\\
    + \textit{AGG$^{T}_{5}$} & 48.8 & \bf{50.8} & 36.4 & 46.9 & 45.7\\
    \bottomrule
  \end{tabularx}
    \caption{56 experiments activating the knowledge in M2M100 with Schedule \textit{I}. }
\label{table:14by4}
\end{table}

Training on $\sim$100 million 
parameters with Geforce RTX 2080 Ti and RTX 3090,
we use a 6-layer encoder and a 6-layer decoder with
512 hidden states, 8 attention heads,
512 word vector size, 2,048 hidden units,
6,000 batch size, 0.1 label smoothing,
2.5 learning learning rate and 1.0 finetuning learning rate, 
0.1 dropout and attention dropout,
a patience of 5 after 190,000 steps in {[N]}$^2$ with an update interval of 1000, 
a patience of 5 for {[N+1]}$^2$ with an update interval of 200, and 
a patience of 25 for {[N+1]} and {[1]}$^2$ with an update interval of 50, 
``adam'' optimizer and
``noam'' decay method \citep{klein2017opennmt, papineni2002bleu}. 

\section{Results}
For simplicity, we use the centeredness method to combine translations 
from all source languages and have one score per metric. To compare across 
different methods, all experiments have the same test set (3,461 lines), 
the intersection of all test sets.

\textbf{\ul{Our models improve over the baselines:}} With Schedule \textit{I}, we observe an average improvement of 24.7 in chrF score over the M2M100 baseline (Table~\ref{table:5by10}). By active learning with 4-gram model, we observe an increase of 28.8 in chrF score over the bilingual baseline. 

\textbf{\ul{Our strategic training schedule improves the translation further by activating the knowledge of M2M100 :}} With Schedule \textit{B} and the 4-gram model, we observe an average improvement of 18.6 in chrF score over the multilingual baseline (Table~\ref{table:4by10}). For Schedule \textit{I}, the increase is 24.8 over the multilingual baseline (Table~\ref{table:5by10}). Indeed, the increase with the activation of M2M100 is greater. 

\subsection{Training Schedules}
We compare 24 training schedules using a randomly sampled seed corpus ($\sim$1,000 lines) to translate into Frisian (Table \ref{table:config2} and \ref{table:config1}). 

\textbf{\ul{Pretraining with {[N]}$^2$ works well without M2M100:}} We compare 8 training schedules without M2M100 (Table~\ref{table:config1}). We find that Schedule \textit{B} (pretraining on {[N]}$^2$ and training on {[N+1]}$^2$ and {[N+1]}) and Schedule \textit{F} (pretraining on {[N]}$^2$ and training on {[N+1]}) work well without M2M100. Schedule \textit{B} gives a chrF score of 51.1 and Schedule \textit{F} gives a chrF score of 51.2.  

M2M100 is useful when a target language and its corresponding source languages are in the M2M100 list and the test set does not overlap with the M2M100 training set. However, we strongly advise discretion, as training data for large pretrained models is usually not clearly specified and most are not trained with endangered languages in mind. M2M100 training data may very likely contain the Bible data, so it only serves as a comparison and provides an alternative view to show that our model is robust with large models. When M2M100 does not apply, our models pretrained with {[N]}$^2$ suffice.

\textbf{\ul{Full stage training increases robustness:}} For models without M2M100 we can use Schedule \textit{B} (Table~\ref{table:14by10}) or F (Table~\ref{table:14by10_old}). Though the results for Frisian are similar, B is much better than F for morphologically rich languages like Pokomchi, Turkmen and Xhosa. Indeed, B with full training is more robust than F, which skips {[N+1]}$^2$. Similarly, for models with M2M100, we can use Schedule \textit{I} (Table~\ref{table:14by4}) or \textit{L} (Table~\ref{table:14by4_old}). Again, Schedule \textit{I} with full training stages perform better than Schedule \textit{L}.  

\textbf{\ul{Applying M2M100 alone gives poor results:}} Schedule \textit{X} produces poor results (Table~\ref{table:config2}). Problems include catastrophic forgetting, bias towards rich resource languages, and unclean data. Existing research shows some released models mislabel their English data as Welsh \citep{radfordrobust}. 

\textbf{\ul{Mixed models with M2M100 perform well:}} A few training schedules beat those pretrained with {[N]}$^2$ (Table~\ref{table:config1}). Schedule \textit{I} (training on 5 stages) gives a chrF score of 52.9, L (training 3 stages skipping {[N+1]} and {[1]}$^2$) gives 52.8, M (training 4 stages skipping {[N+1]}$^2$) gives 52.7, J (training 4 stages skipping {[1]}$^2$) gives 51.8, and N (training 3 stages skipping {[N+1]}$^2$ and {[1]}$^2$) gives 51.9. All are higher than 
those without M2M100. 

\textbf{\ul{Adapting M2M100 to the domain and then to the endangered language works best:}} Schedule \textit{I} (training on 5 stages) with score 52.9 performs best. These models first adapt M2M100 to the domain by doing another pretraining on N$^2$. After adapting M2M100 to the domain, we adapt the model to the endangered language by training on {[N+1]}$^2$. The final two stages {[N+1]} and {[1]}$^2$ are optional.  

\subsection{Active Learning Methods}

Using Schedule \textit{B} without M2M100, and \textit{L} with M2M100, we compare 14 active learning methods across languages (Table~\ref{table:14by10} and \ref{table:14by4}). 

\textbf{\ul{Normalizing by sequence length improves density:}} Without normalization, the model chooses longer sentences with many rare words. Normalization improves density. For Sesotho, the chrF score is 39.0 without normalization and 41.6 with it. 

\textbf{\ul{Marginal benefit of increasing n-gram order wanes:}} Existing research shows bigrams suffice \citep{eck2008developing}. As the n-gram order increases, the data gets sparser and the marginal benefit 
subsides. Hmong has the best score (46.1) using bigrams.

\textbf{\ul{Tipping points vary with language:}} The optimal highest n-gram order may differ from language to language. 4-grams work best for Frisian while bigrams work best for Hmong. Hmong is an isolating language while Frisian is a fusional language. A possible explanation is that higher n-grams may have more impact on fusional languages.

\textbf{\ul{Entropy and n-gram methods both beat baselines and higher n-gram models perform best:}} KenLM is much faster and performs better than NLTK. The entropy method using KenLM beats both baselines. Frisian has a chrF score of 52.7 with the entropy method using KenLM. This is much higher than the baselines: \textit{Luke} (47.5) and \textit{Rand} (50.5). The 4-gram model (53.6) is higher because building LMs from a few lines of data may not be accurate. Simpler n-gram models work better than more evolved entropy models with small data. 

\textbf{\ul{Aggregation over all languages serves as a universal ranking:}}  The first 10 active learning methods are based on learning from one reference language and generalizing to the endangered language, while the last 4 focus on aggregation over multiple languages (Table~\ref{table:14by10} and \ref{table:14by4}). 
For Welsh, aggregation over multiple languages (48.2 with most spoken languages) performs better than those that rely on one reference language; but for other languages aggregation performs worse. 
Aggregation over all languages performs better than other aggregation methods for all languages except Welsh. 
This hinges on the reference language. For Frisian, choosing English (a Germanic language) as a reference language, performs better than aggregation. For Welsh (a Celtic language), choosing a reference language that is not as close, performs worse. But we often do not have such information for endangered languages. In such cases, universal ranking by aggregating over all languages is useful. 


\textbf{\ul{Our active learning methods mimic curriculum learning:}} Our models pick short and simple sentences first, 
emulating curriculum learning and helping human translators \citep{bengio2009curriculum, graves2017automated, jiang2015self}. 

\textbf{\ul{All active learning methods cover different genres:}} Our methods pick a mix of sentences from different genres, sentence
lengths and complexity levels. Moreover, our methods pick narrative sentences first, which is helpful for human translators.


\textbf{\ul{Our model captures some language subtleties:}} Apart from the metrics, we showed our translation to native speakers (Table \ref{table:quali_srclanguages}). We translate "He sees that it is good" to "lug ca rua huv nwg lu sab" ("He puts it in the liver") in Hmong, which uses liver to express joy. This increases lexical choice. 

\textbf{\ul{Our models and mixed models perform better than M2M100 alone:}} M2M100 often produces extremely short sentences or repetition. Our models do not have those issues. 

\section{Future Work}
We propose 24 training schedules for translation into endangered languages. We also propose and compare 14 active learning methods to build seed corpus without any endangered language data. Our model is robust with large multilingual models. 

While the industry trend is to move towards bigger models with bigger data, our minimalist approach not only uses fewer languages, but we also aggregate over fewer languages. 
This saves computation power and resources, and therefore time and money, while improving translation performance. 

However, we still face challenges with the lack of local coherence and context.
The excerpt-based approach
enjoys advantage with formality, cohesion and contextual
relevance. Active learning methods, on the contrary,
do not have consecutive sentences and therefore lose local coherence and pose challenges to human
translators \citep{muntes2012context, denkowski2015machine, sperber2017transcribing, maruf2019survey, webster2020gutenberg, zhou2021active, salunkhe2016hybrid}. 
This is an active research area.

Evaluation is still a challenge. It is difficult to find native
speakers and establish 
long-term collaborations. There is also much variety among endangered languages. Some are
more accessible than others and these might provide earlier, realistic
evaluation of our method. 
Empowering endangered languages is not just a technology
problem. It requires much efforts in communication 
with local communities. 
Through our technologies, we would like to work with local communities to revive endangered languages 
and bring them to flourish. 

\section*{Acknowledgements}
Thanks to Alan Black, Alon Lavie, Graham Neubig, Uri Alon, David Mortensen, Kevin Haworth, Christian Hallstein for the great discussions and suggestions. 

\bibliographystyle{acl_natbib}
\bibliography{thesis}

\clearpage
\appendix

\section{Appendices}

\begin{table}[b!]
  \small
  \centering
  \begin{tabularx}{\columnwidth}{p{1.2cm}p{0.7cm}p{0.7cm}p{1.1cm}p{0.7cm}p{0.8cm}} 
    \toprule
    $\uparrow$chrF & Frisian & Welsh & Hungarian & Spanish  & Average \\
    \midrule
    \multicolumn{6}{l}{\textbf{Baselines:} }\\
    \textit{Luke} & 49.1 & 41.7 & 38.3 & 48.7 & 44.5 \\
    \textit{Rand} & 52.8 & 46.8 & 41.9 & 52.9 & 48.6 \\
    \midrule
    \multicolumn{6}{l}{\textbf{Our Models:} }\\
    \textit{S} & 51.6 & 44.8 & 40.7 & 52.0 & 47.3 \\
    \textit{SN} & 53.2 & 45.8 & 42.2 & 52.9 & 48.5 \\
    \textit{SNG$_{2}$} & 54.2 & 47.6 & 42.5 & 53.8 & 49.5 \\
    \textit{SNG$_{3}$} & 53.7 & 47.9 & \bf{43.3} & \bf{54.5} & 49.9 \\
    \textit{SNG$_{4}$} & \bf{54.3} & 48.5 & 43.2 & 54.4 & \bf{50.1} \\
    \textit{SNG$_{5}$} & 53.9 & 48.6 & 43.2 & 54.5 & 50.1 \\
    \textit{ENT$^{N}$} & 52.1 & 44.8 & 40.7 & 52.4 & 47.5 \\
    \textit{ENT$^{K}$} & 53.7 & 46.7 & 43.1 & 53.7 & 49.3 \\
    \textit{AGG$^{A}_{5}$} & 48.4 & 43.2 & 37.1 & 48.4 & 44.3 \\
    \textit{AGG$^{S}_{5}$} & 47.3 & 48.1 & 36.1 & 47.1 & 44.7 \\
    \textit{AGG$^{M}_{5}$} & 46.9 & 47.8 & 36.3 & 47.2 & 44.6 \\
    \textit{AGG$^{T}_{5}$} & 47.1 & \bf{48.8} & 36.1 & 46.8 & 44.7 \\
    \bottomrule
  \end{tabularx}
    \caption{56 experiments integrated with M2M100 on Schedule \textit{L}. }
\label{table:14by4_old}
\end{table}

\begin{table*}[b]
  \small
  \centering
  \begin{tabularx}{\textwidth}{p{1.2cm}p{0.8cm}p{0.8cm}p{1.1cm}p{0.9cm}p{0.9cm}p{0.7cm}p{0.7cm}p{1.1cm}p{1.1cm}p{0.8cm}p{0.9cm}} 
    \toprule
   $\uparrow$chrF & Frisian & Hmong & Pokomchi & Turkmen & Sesotho & Welsh & Xhosa & Indonesian & Hungarian & Spanish & Average \\
    \midrule
    \multicolumn{12}{l}{\textbf{Baselines:} }\\ 
    \textit{Luke} & 47.5 & 38.2 & 37.4 & 33.8 & 38.5 & 38.5 & 29.2 & 41.7 & 31.5 & 46.3 & 38.3\\
    \textit{Rand} & 51.3 & 38.9 & 41.5 & 36.4 & 39.0 & 43.1 & 32.1 & 45.3 & 34.8 & 50.2 & 41.3 \\
    \midrule 
    \multicolumn{12}{l}{\textbf{Our Models:} }\\
    \textit{S} & 48.7 & 35.8 & 39.8 & 27.6 & 36.1 & 38.1 & 29.4 & 41.5 & 32.5 & 47.5 & 37.7 \\
    \textit{SN} & 50.9 & 38.4 & 41.5 & 36.9 & 38.7 & 41.1 & 32.5 & 44.8 & 33.1 & 49.2 & 40.7 \\
    \textit{SNG$_{2}$} & 52.9 & 40.9 & 42.4 & 37.3 & 41.0 & 44.3 & 33.4 & 45.8 & 35.8 & 51.2 & 42.5 \\
    \textit{SNG$_{3}$} & 53.1 & 41.8 & \bf{43.2} & \bf{38.4} & 41.9 & 45.6 & \bf{34.0} & 47.0 & 36.4 & 52.2 & \bf{43.4} \\
    \textit{SNG$_{4}$} & \bf{53.6} & 41.8 & 42.2 & 38.1 & 41.7 & 44.5 & 33.5 & \bf{47.5} & 36.7 & \bf{52.5} & 43.2 \\
    \textit{SNG$_{5}$} & 53.0 & 41.5 & 42.0 & 38.1 & \bf{42.3} & 45.1 & 33.5 & 47.3 & 36.4 & 52.2 & 43.1 \\
    \textit{ENT$^{N}$} & 50.7 & 39.5 & 34.0 & 34.8 & 39.4 & 42.5 & 32.4 & 44.4 & 33.9 & 48.6 & 40.0 \\
    \textit{ENT$^{K}$} & 52.5 & \bf{42.4} & 42.3 & 38.5 & 41.6 & 43.4 & 33.6 & 47.1 & \bf{37.1} & 51.7 & 43.0 \\
    \textit{AGG$^{L}_{5}$} & 47.4 & 38.8 & 38.9 & 33.2 & 37.3 & 40.1 & 28.9 & 41.6 & 31.7 & 45.7 & 38.4 \\
    \textit{AGG$^{F}_{5}$} & 44.6 & 36.0 & 37.1 & 30.9 & 35.8 & 44.3 & 27.8 & 39.2 & 30.7 & 43.9 & 37.0 \\
    \textit{AGG$^{P}_{5}$} & 45.2 & 36.6 & 36.9 & 30.8 & 35.6 & 44.9 & 27.9 & 39.0 & 30.5 & 43.8 & 37.1 \\
    \textit{AGG$^{N}_{5}$} & 45.4 & 36.8 & 37.1 & 31.3 & 35.7 & \bf{46.0} & 28.0 & 39.2 & 30.2 & 43.8 & 37.4 \\
    \bottomrule
  \end{tabularx}
    \caption{140 experiments comparing 14 active learning methods translating into 10 different languages on Schedule \textit{F}.  
    }
\label{table:14by10_old}
\end{table*}

For simplicity, in Table \ref{table:ethnologue_languages} 
Pokomchi is Eastern Pokomchi, Hmong is Hmong Hoa, 
Kanjobal is Eastern Kanjobal, Mam is Northern Mam, 
Cuzco is Cuzco Quechua, Ayacucho is Ayacucho Quechua, 
Bolivian is South Bolivian Quechua, and Huallaga is
Huallaga Quechua, Chinese is Traditional Chinese, 
Haka is Haka Chin, Siyin is Siyin Chin, Falam is 
Falam Chin, Kpelle is Kpelle Guinea. 

In Table~\ref{table:4by10}, our model with training scheduling uses Schedule \textit{B}, our model with active learning uses \textit{SNG$_{4}$}.  In Table~\ref{table:5by10}, our model with training scheduling uses Schedule \textit{I}, our model with active learning uses \textit{SNG$_{4}$}.  

In the entropy score function in Table~\ref{table:ngramScore}, we use highest n-gram order of 2 for NLTK's LM, we use highest n-gram order of 2 for the two halves ($H_{l}^K$ and $H_{r}^K$) and order of 5 for the sampled data ($H_{c}^K$) for KenLM. Since KenLM needs at least a few words to
start with, we use MLE as a warm start to select up to
5 sentences before launching KenLM. 

For finetuning from a M2M100 Model, 
training on $\sim$418 million 
parameters with Geforce RTX 3090,
we use a 12-layer encoder and a 12-layer decoder with
1024 hidden states, 16 attention heads,
1024 word vector size, 4,096 hidden units, 
0.2 label smoothing,
0.0002 training learning rate and finetuning 0.00005 learning rate, 
0.1 dropout and attention dropout,
``adam'' optimizer and
``noam'' decay method \citep{fan2021beyond, schwenk2019ccmatrix, el2019massive}. 

\begin{table*}[t]
  \small
  \centering
  \begin{tabularx}{\textwidth}{p{1.8cm}p{0.7cm}p{0.7cm}p{1.1cm}p{0.9cm}p{0.8cm}p{0.6cm}p{0.6cm}p{1.0cm}p{1.1cm}p{0.8cm}p{0.9cm}} 
   \toprule
   Seed Corpus Size & Frisian & Hmong & Pokomchi & Turkmen & Sesotho & Welsh & Xhosa & Indonesian & Hungarian & Spanish & Average \\
    \midrule
    Word count & 25695 & 31249 & 36763 & 17354 & 25642 & 25786 & 15017 & 22318 & 18619 & 22831 & 24127\\
    \midrule
    \multicolumn{6}{l}{Line count for each experiment}\\
    \multicolumn{6}{l}{\textbf{Baselines:} }\\
    \textit{Luke} & 1151 & 1151 & 1151 & 1151 & 1151 & 1151 & 1151 & 1151 & 1151 & 1151 & 1151\\
    \textit{Rand} & 1022 & 1001 & 1101 & 1045 & 976 & 1117 & 988 & 1065 & 1066 & 1023 & 1040 \\
     \midrule
    \multicolumn{6}{l}{\textbf{Our Models:} }\\
    \textit{S} & 692 & 654 & 832 & 689 & 657 & 771 & 598 & 634 & 644 & 682 & 685 \\
    \textit{SN} & 1522 & 1399 & 1522 & 1524 & 1434 & 1595 & 1501 & 1601 & 1545 & 1488 & 1513 \\
    \textit{SNG$_{2}$} & 1484 & 1350 & 1490 & 1454 & 1369 & 1557 & 1418 & 1513 & 1468 & 1463 & 1457 \\
    \textit{SNG$_{3}$} & 1385 & 1319 & 1468 & 1416 & 1317 & 1439 & 1368 & 1451 & 1415 & 1365 & 1394 \\
    \textit{SNG$_{4}$} & 1327 & 1295 & 1419 & 1367 & 1279 & 1409 & 1309 & 1426 & 1374 & 1310 & 1352\\
    \textit{SNG$_{5}$} & 1289 & 1289 & 1397 & 1311 & 1280 & 1381 & 1256 & 1359 & 1334 & 1273 & 1317 \\
    \textit{ENT$^{N}$} & 1796 & 1721 & 1769 & 1840 & 1761 & 1914 & 1839 & 1967 & 1884 & 1805 & 1830\\
    \textit{ENT$^{K}$} & 1340 & 1287 & 1507 & 1266 & 1132 & 1405 & 1128 & 1358 & 1264 & 1327 & 1301 \\
    \textit{AGG$^{A}_{5}$} & 984 & 1025 & 1060 & 998 & 967 & 1031 & 1016 & 1018 & 993 & 958 & 1005 \\
    \textit{AGG$^{S}_{5}$} & 1049 & 1084 & 1152 & 1043 & 1025 & 1182 & 1147 & 1093 & 1076 & 1019 & 1087\\
    \textit{AGG$^{M}_{5}$} & 1058 & 1097 & 1159 & 1109 & 1025 & 1232 & 1159 & 1101 & 1087 & 1018 & 1105\\
    \textit{AGG$^{T}_{5}$} & 1048 & 1094 & 1153 & 1101 & 1020 & 1274 & 1141 & 1101 & 1087 & 1014 & 1103\\
    \bottomrule
  \end{tabularx}
    \caption{Seed Corpus Size for different target languages. The seed corpus gives rise to both training data and validation data, therefore the training size is smaller than the above. Note that all experiments for a given target language share the same number of words, although they have different number of lines. Since each language use different number of words to express the same meaning of a given text, we choose the number of words in the given book "Luke" as the standard reference for each target language. For example, "Luke" in Xhosa contains 15,017 words while "Luke" in Frisian contains 25,695 words. }
\label{table:datasize}
\end{table*}

\begin{table*}[t]
  \small
  \centering
  \begin{tabularx}{\textwidth}{p{1.2cm}p{6cm}p{7.8cm}} 
    \toprule
    Target & System Translation & Reference \\
    \midrule    
    Frisian & mar Ruth sei: Ik scil dy net forlitte, en ik scil fen dy net weromkomme; hwent hwer "tstû hinnegeane, den scil ik hinnegean, en dêr scil ik dy fornachtsje. dyn folk is myn folk, en dyn God is myn God. & mar Ruth sei: Sit net tsjin my oan, dat ik jo forlitte en weromtsjen scil; hwent hwer "t jo hinne geane, dêr scil ik hinne gean, en hwer "t jo fornachtsje, dêr scil ik fornachtsje; jins folk is myn folk en jins God is myn God;\\
    Hmong & Lauj has rua nwg tas, "Tsw xob ua le ntawd, kuv yuav moog rua koj lub chaw kws koj moog, hab kuv yuav nyob huv koj haiv tuabneeg. koj yog kuv tug Vaajtswv." & tassws Luv has tas, "Tsw xob has kuas kuv tso koj tseg ncaim koj rov qaab moog. koj moog hovtwg los kuv yuav moog hab, koj nyob hovtwg los kuv yuav nyob hov ntawd hab, koj haiv tuabneeg los yog kuv haiv tuabneeg hab, koj tug Vaajtswv los yog kuv tug Vaajtswv. \\
    Pokomchi & eh je' wili i xq'orarik reh i Rut: Maacanaa' chih taj i hin. re' hin naa nub'anam aweh chupaam i ye'aab' naa nuk'achariik ayu'. re' hin naa nuk'achariik awuuk', eh re' hin naa nukahniik chi nuDios, inki. & re' Rut je' wili i chaq'wik xub'an: Maa pahqaaj aakuyariik weh re' hin ma' jaruuj nee tinukanaa' kahnoq, xa aha' pa' nee tiooj i hat, nee wo' kinooj chawiij, xa aha' pa' nee ti k'achariik i hat ar nee kink'acharik i hin. eh re' aatinamiit re' wo' re' nutinamiit i hin, eh re' aaDios re' wo' re' nuDios i hin. \\
    Turkmen & Rut: oňa: "Sen nirä gitseň, men hem seniň ýanyňa gitmerin. Sen nirä gitseň, men hem seniň halkym bolaryn. Men seniň Hudaýym bolaryn. & emma Rut: "Seni terk edip ýanyňdan gitmegi menden haýyş etme. sen Nirä gitseň, Menem şol ýere gitjek. sen nirede bolsaň, Menem şol ýerde boljak. seniň halkyň - meniň halkym, seniň Hudaýyň meniň Hudaýym bolar. \\
    Sesotho & yaba Ruthe o re ho yena: "O se ke wa tloha ho wena, hobane ke tla ya le wena, ke tla ya le wena, mme ke tla ya hona moo. setjhaba sa ka, le Modimo wa hao." & empa Ruthe a re: "O se ke wa nqobella hore ke kgaohane le wena, kapa hore ke se ke ka tsamaya le wena, hobane" moo o yang teng ke tla ya teng, moo o phelang teng ke tla phela teng; tjhaba sa heno e be tjhaba sa heso, Modimo wa hao e be Modimo wa ka. \\    
    Welsh & a Ruth a ddywedodd, Nuw gael arnaf fi, atolwg, atolwg, oddi wrthyt: canys lle yr wyt yn myned, ac yno yr wyt yn myned, y byddaf fy hun. dy bobl yw fy bobl, a'th Dduw yw fy Duw. & a Ruth a ddywedodd, Nac erfyn arnaf fi ymado â thi, i gilio oddi ar dy ôl di: canys pa le bynnag yr elych di, yr af finnau; ac ym mha le bynnag y lletyech di, y lletyaf finnau: dy bobl di fydd fy mhobl i, a'th Dduw di fy Nuw innau:\\
    Xhosa & URute waphendula wathi: "Undiyekeli ukuba ndixhamle, kuba ndiza kuhlala apho uthanda khona. mna ndiza kuba ngabantu bam, abe nguThixo wam." & Waphendula uRute wathi: "Sukundinyanzela usithi mandikushiye. apho uya khona, nam ndiya kuya, ndiye kuhlala nalapho uhlala khona, amawenu abe ngamawethu, noThixo wakho abe nguThixo wam. \\
    Indonesian & tetapi Rut: menjawab: "Janganlah engkau meninggalkan aku dan pulang ke tempat kediamanmu, sebab aku akan pergi dan berdiam di mana engkau diam, sebab orang-orangmu akan menjadi umat-Ku dan Allahmu." & tetapi kata Rut: "Janganlah desak aku meninggalkan engkau dan pulang dengan tidak mengikuti engkau; sebab ke mana engkau pergi, ke situ jugalah aku pergi, dan di mana engkau bermalam, di situ jugalah aku bermalam: bangsamulah bangsaku dan Allahmulah Allahku; \\
    Hungarian & Ruth így felelt: Nem kérlek téged, hogy gondolj meg téged, mert csak hozzád megyek, és én otthagytam, hogy legyenek hozzád. a te népem az én, és az én Istenem az én. & de Ruth azt felelte: Ne unszolj engem, hogy elhagyjalak és visszatérjek tőled. mert ahová te mégy, odamegyek, ahol te megszállsz, ott szállok meg. Néped az én népem, és Istened az én Istenem. \\
    Spanish & y Rut: dijo a David: No me permite de ti, y me quitaré de ti; porque donde vayas, yo iré a donde vayas, y habitaré; y tu pueblo es mi pueblo, y tu Dios es mi Dios. & respondió Rut: No me ruegues que te deje, y me aparte de ti; porque a dondequiera que tú fueres, iré yo, y dondequiera que vivieres, viviré. tu pueblo será mi pueblo, y tu Dios mi Dios. \\
    \bottomrule
  \end{tabularx}
    \caption{Qualitative evaluation using \textit{SNG$_{5}$} 
    to translate into each target language. }
\label{table:quali_srclanguages}
\end{table*}

\end{document}